\documentclass{ws-ijns}
\usepackage{graphicx}
\usepackage{hyperref}
\usepackage[english]{babel}
\usepackage{soul}
\usepackage{xcolor}
\sethlcolor{yellow} 

\usepackage[super,sort,compress]{cite}

\begin{document}

\catchline{0}{0}{2005}{}{}

\markboth{T. Tsiolakis et al.}{Evaluation of Bio-Inspired Models under Different Learning Settings For Energy Efficiency in Network Traffic Prediction}

\title{Evaluation of Bio-Inspired Models under Different Learning Settings For Energy Efficiency in Network Traffic Prediction.}

\author{Theodoros Tsiolakis}

\address{Department of Electrical and Computer Engineering, Democritus University of Thrace\\
Xanthi, 67100, Greece\\
E-mail: ttsiolak@ee.duth.gr\\
www.duth.gr}

\author{Nikolaos Pavlidis}
\address{Department of Electrical and Computer Engineering, Democritus University of Thrace\\
Xanthi, 67100, Greece\\}

\author{Vasileios Perifanis}
\address{Department of Electrical and Computer Engineering, Democritus University of Thrace\\
Xanthi, 67100, Greece\\}

\author{Pavlos Efraimidis}
\address{Department of Electrical and Computer Engineering, Democritus University of Thrace\\
Xanthi, 67100, Greece\\}

\maketitle

\begin{abstract}
Cellular traffic forecasting is a critical task that enables network operators to efficiently allocate resources and address anomalies in rapidly evolving environments. The exponential growth of data collected from base stations poses significant challenges to processing and analysis. While machine learning (ML) algorithms have emerged as powerful tools for handling these large datasets and providing accurate predictions, their environmental impact, particularly in terms of energy consumption, is often overlooked in favor of their predictive capabilities.
This study investigates the potential of two bio-inspired models: Spiking Neural Networks (SNNs) and Reservoir Computing through Echo State Networks (ESNs) for cellular traffic forecasting. The evaluation focuses on both their predictive performance and energy efficiency. These models are implemented in both centralized and federated settings to analyze their effectiveness and energy consumption in decentralized systems. Additionally, we compare bio-inspired models with traditional architectures, such as Convolutional Neural Networks (CNNs) and Multi-Layer Perceptrons (MLPs), to provide a comprehensive evaluation. Using data collected from three diverse locations in Barcelona, Spain, we examine the trade-offs between predictive accuracy and energy demands across these approaches.
The results indicate that bio-inspired models, such as SNNs and ESNs, can achieve significant energy savings while maintaining predictive accuracy comparable to traditional architectures. Furthermore, federated implementations were tested to evaluate their energy efficiency in decentralized settings compared to centralized systems, particularly in combination with bio-inspired models. These findings offer valuable insights into the potential of bio-inspired models for sustainable and privacy-preserving cellular traffic forecasting.
\end{abstract}


\begin{multicols}{2}
\section{Introduction}
Accurately predicting network traffic and load is essential for improving network management and ensuring efficient operation for telecommunication providers. Reliable forecasting enables proactive resource allocation, effective capacity planning, and improved service quality, especially in high-demand scenarios. The primary challenge lies in maintaining robust connectivity for all devices while avoiding unnecessary resource wastage. Machine learning (ML) algorithms have become indispensable tools for addressing this challenge, offering capabilities such as real-time decision-making, anomaly detection, failure prediction, and dynamic spectrum management. These features make ML a critical component for creating smarter, more sustainable telecommunication networks~\cite{Miozzo2021Distributed, pavani2023machine}.

This study is motivated by the rapid growth of data generated by telecommunication systems, which is driven by the increasing use of connected devices. The large amounts of data involved require sophisticated predictive models that need significant computational resources to process. As a result, the training of these models often consumes substantial amounts of energy, time, and storage, creating challenges for both efficiency and sustainability. This underscores the need for approaches that balance strong predictive performance with reduced energy usage. This consideration is particularly important in the context of federated learning, where models are trained on edge devices with limited computational and energy resources. In such scenarios, it is critical to identify models that can be effectively trained under constrained conditions. 

Furthermore, the availability of models with low power and computational requirements is essential for ensuring service continuity during emergencies—such as power outages or infrastructure disruptions—where maintaining operation at reduced capacity is vital.

In this work, we focus on identifying and analyzing the trade-offs between predictive performance and power consumption across different model architectures and deployment settings.



In this study, we explore the use of bioinspired SNNs and Reservoir Computing (RC) with Echo State Networks (ESNs) for cellular traffic forecasting. We examine several types of Leaky Integrate-and-Fire (LIF) neurons in SNNs, including Lapicque, Leaky, RLeaky, Synaptic, and Alpha neurons~\cite{eshraghian2021snntorch}, as well as ESNs~\cite{Jaeger2001RC_Original}. These models are compared with traditional machine learning methods, such as Multi-Layer Perceptrons (MLPs) and Convolutional Neural Networks (CNNs). Both centralized and federated learning settings are used in the evaluation to account for modern distributed computation approaches. Federated learning, in particular, reduces the need for centralized data storage, offering a more energy-efficient way to train models and we aim to evaluate the trade-offs between predictive accuracy and energy efficiency. To achieve this, we use a Sustainability Index that combines key metrics like power consumption, prediction error, and a size parameter to provide a comprehensive evaluation of model efficiency.

To the best of our knowledge, this is the first study to compare the energy consumption of bio-inspired and traditional models for cellular traffic forecasting, evaluated across both centralized and federated learning settings.

We further explore the power efficiency of federated learning, which was initially introduced to enhance privacy in machine learning. Recent studies~\cite{Wang2019FLEnergy} indicate that federated learning can also reduce power consumption, offering a dual benefit of improved privacy and energy efficiency. This makes it a promising framework for sustainable and privacy-preserving machine learning.

The results aim to guide researchers and practitioners toward developing machine learning models that balance high performance with lower energy use. The rest of the paper is structured as follows: Section~\ref{sec:related_work} reviews related work on Spiking Neural Networks (SNNs), Reservoir Computing (RC), and Federated Learning. Section~\ref{sec:system_model} describes the dataset and research problem, along with the models and evaluation methods. Section~\ref{sec:experiments} explains the experimental setup and methodology. Section~\ref{sec:evaluation} presents the results and analysis. Section~\ref{sec:discussion} discusses the findings, challenges, and future directions. Finally, Section~\ref{sec:conclusion} concludes by summarizing the key outcomes and contributions of the study.

\section{Related Work} \label{sec:related_work}

\subsection{Traffic Prediction}
Recent advancements in cellular traffic forecasting have demonstrated the effectiveness of machine learning and federated learning approaches in managing 5G base stations. Perifanis et al. (2023)~\cite{perifanis20235fed5g} introduced a federated learning framework for traffic prediction in 5G base stations, highlighting the model’s ability to preserve data privacy while enabling distributed training across multiple stations. Their work emphasized the importance of efficient time-series forecasting to optimize resource allocation and reduce energy consumption in large-scale cellular networks.

Building on this foundation, Pavlidis et al. (2024)~\cite{Pavlidis2024} extended the concept by applying federated models to predict traffic across a larger number of base stations, while also investigating the impact of fine tuning several experimental parameters. They demonstrated how federated learning could improve forecasting accuracy and reduce unnecessary communication overheads.

Zhao et al. (2023) proposed the FL-DeepAR framework, which uses deep autoregressive networks for federated traffic prediction. This approach addressed statistical heterogeneity and incorporated advanced gradient-based aggregation techniques, achieving superior prediction performance on real-world datasets~\cite{zhao2023research}.

Similarly, Guesmi et al. (2024) ~\cite{Guesmi2024} proposed advanced predictive modeling techniques to enhance traffic forecasting in emerging cellular networks. Their research leveraged a combination of deep learning and federated learning algorithms to address the challenges of time-series forecasting in dynamic and heterogeneous environments, further reinforcing the potential of distributed machine learning in telecommunications.

Gao et al. (2024) developed FedCE, a communication-efficient federated learning framework that reduced communication overhead while improving prediction accuracy using gradient compression and adaptive aggregation strategies. The results highlighted significant improvements in prediction accuracy with optimized resource utilization~\cite{gao2024communication}.

Additionally, Lee et al. (2024) proposed a personalized federated learning approach for mobile traffic prediction, which utilized layer-wise aggregation and adaptive layer freezing to improve prediction accuracy and communication efficiency. This method demonstrated its potential for reducing overhead and ensuring scalability in dynamic network environments~\cite{lee2024layer}.

Lastly, Kalam et al. (2023) introduced a federated learning approach for 5G traffic forecasting, demonstrating significant improvements in accuracy over traditional centralized methods while ensuring privacy preservation across distributed networks~\cite{kalam20235g}.

Collectively, these studies underscore the transformative potential of federated learning and complementary machine learning approaches in cellular traffic forecasting. They offer scalable, privacy-preserving, and resource-efficient solutions to optimize the performance and sustainability of 5G and beyond networks, paving the way for the next generation of mobile communication systems.

\subsection{SNN in Prediction Tasks}
Recent advancements in spiking neural networks (SNNs) demonstrate their potential for efficient and accurate network traffic prediction, particularly in energy-constrained environments. In our previous work ~\cite{Tsiolakis2024} we explored SNNs in the context of cellular traffic forecasting, emphasizing their ability to reduce power consumption compared to conventional machine learning models. By leveraging the bio-inspired architecture of SNNs, the study addressed the growing demand for environmentally sustainable machine learning solutions in telecommunications.

Similarly, Kang et al. (2024) ~\cite{KangHotspotPrediction2024} investigated the use of SNNs in neuromorphic systems for traffic hotspot prediction in network-on-chip (NoC) architectures. The findings showcased how SNN-based approaches not only improve computational efficiency but also enhance routing strategies, contributing to a more energy-efficient processing pipeline.

Early work by Ghosh-Dastidar and Adeli (2009) ~\cite{GHOSHDASTIDAR2009} introduced a novel supervised learning algorithm for multiple spiking neural networks, demonstrating their predictive capabilities in the clinically significant task of epilepsy and seizure detection. This study highlights the potential of SNNs for time-sensitive and pattern recognition tasks, laying groundwork for later applications in domains such as image classification and reservoir computing.

Nguyen et al. (2023) ~\cite{sanaullah2023} evaluated LIF-based spiking neural networks for image classification using the RAVSim simulator, demonstrating their predictive potential in structured visual tasks.

Together, these studies underscore the transformative potential of SNNs in network traffic prediction, providing scalable, low-power solutions for real-time applications while maintaining high predictive accuracy.

\subsection{RC in Traffic Prediction}
Reservoir computing has emerged as a promising framework for cellular traffic forecasting due to its ability to process complex temporal data efficiently while minimizing computational power consumption. Li et al. (2021) ~\cite{Li2021} introduced a deep multi-reservoir regression learning network tailored for cellular traffic prediction in multi-access edge computing environments. This innovative approach leverages the strengths of reservoir computing to provide real-time and accurate predictions, demonstrating its scalability and efficiency in handling dynamic traffic patterns while maintaining low energy overheads.

Further advancing the field, Zhang and Vargas (2023) ~\cite{Zhang2023SurveyRC} applied reservoir computing in the context of 5G and beyond networks. Their study emphasized how reservoir computing architectures enhance forecasting accuracy while addressing sustainability challenges in mobile networks. By leveraging innovative learning techniques, their work provides actionable insights for integrating reservoir computing frameworks into modern telecommunications infrastructures, ensuring both predictive performance and energy efficiency.
Xue et al. (2021) ~\cite{XueYu2021} proposed a self-adaptive PSO-optimized Echo State Network for time series prediction, highlighting the potential of reservoir computing in modeling complex temporal patterns efficiently.
Collectively, these studies underline reservoir computing's transformative potential for cellular traffic forecasting, particularly in energy-constrained and dynamic environments like 5G networks. Its ability to balance computational efficiency with prediction accuracy makes it a critical tool for future network management and optimization.

\subsection{Federated Bio Inspired in Traffic Prediction}
Federated Learning (FL) addresses privacy concerns in mobile networks but faces challenges like high energy consumption, communication overhead, and convergence issues. Aouedi et al. (2023) \cite{aouedi2023hfedsnn} introduced HFedSNN, a Hierarchical Federated Learning (HFL) model using Spiking Neural Networks (SNNs) to overcome these issues. HFedSNN reduces energy consumption by 4.3x, communication overhead by 26x, and improves accuracy by 4.48\% compared to FL with ANNs (FedANN), making it highly efficient for non-IID data scenarios.

Similarly, Bacciu et al. (2021) \cite{Bacciu2021} introduced Federated Reservoir Computing (FRC) that leverages bio-inspired models like Echo State Networks (ESNs) for low-power, distributed temporal data processing. ESNs excel in handling time-series data while minimizing energy use, offering scalable, privacy-preserving solutions. 

Together, HFL with SNNs and FRC provide a powerful framework for robust, energy-efficient, and scalable AI in edge computing environments.

Our work advances the field of cellular traffic prediction by addressing critical challenges in energy efficiency and sustainability, particularly in the context of bioinspired models. While prior research has demonstrated the potential of federated learning to improve prediction accuracy and preserve privacy, much of the focus has been on traditional machine learning methods. Similarly, bioinspired models, such as SNNs and RC, have been explored for their energy efficiency in specific use cases, but their integration with federated learning frameworks remains underexplored. Existing approaches often overlook the trade-offs between energy consumption, predictive performance, and environmental impact.

Our study distinguishes itself by combining SNNs and RC models within federated learning environments to assess their predictive capacity and energy efficiency. Unlike prior studies that prioritize either centralized configurations or accuracy alone, we employ a Sustainability Index (See Section \ref{sec:system_model}) to evaluate the trade-offs across multiple dimensions, including power consumption, prediction error, and data size. This comprehensive approach, coupled with comparative evaluations against traditional architectures such as MLPs and CNNs, provides deeper insights into how bioinspired models can support sustainable and scalable solutions for cellular traffic forecasting. By leveraging real-world datasets and consistent experimental setups, our work offers a holistic framework for advancing energy-efficient and privacy-preserving network management practices.

\section{System Model} \label{sec:system_model}
This section outlines the framework used to evaluate the performance and energy efficiency of the different models and settings. This includes details about the dataset, the problem formulation, and the model architectures implemented in both centralized and federated learning setups.

\subsection{Dataset}
In our analysis, we utilized a real-world dataset with LTE traffic measurements that was gathered from three different areas in Barcelona, Spain. The dataset was collected under SUPERCOM initiative,\footnote{\href{https://supercom.cttc.es/}{https://supercom.cttc.es/}} from Downlink Control Information (DCI) messages, which are transmitted through the PDCCH every Transmission Time Interval (TTI). Then, the raw data are aggregated into two minutes intervals. Each measurement contains eleven features regarding network traffic (see Table ~\ref{tab:dataset_features}) and the downstream task is to predict the five of them, i.e. \textit{RNTI Count}, \textit{UpLink}, \textit{DownLink}, \textit{RB Up} and \textit{RB Down}, using historical measurements for the next timestep.

\begin{table*}
\tbl{List of dataset features. \label{tab:dataset_features}}
{\begin{tabular}{clc}
\toprule
Feature      & Description                                                      & Target Column \\ \hline
DownLink     & The downlink transport block size in bits                        & \checkmark             \\
UpLink       & The uplink transport block size in bits                          & \checkmark             \\
RNTI Count   & The average RNTI counter, indicating the average number of users & \checkmark             \\
RB Up        & The average number of allocated resource blocks in the UL        & \checkmark             \\
RB Down      & The average number of allocated resource blocks in the DL        & \checkmark             \\
RB Up Var    & The normalized variance of RB Up                                 &               \\
RB Down Var  & The normalized variance of RB Down                               &               \\
MCS Up       & The average Modulation and Coding Scheme (MCS) index in the UL   &               \\
MCS Down     & The average Modulation and Coding Scheme (MCS) index in the DL   &               \\
MCS Up Var   & The normalized variance of MCS Up                                &               \\
MCS Down Var & The normalized variance of MCS Down                              &               \\ 
\botrule
\end{tabular}}
\end{table*}

Specifically, the measurements are gathered in the following locations:
\begin{itemize}
    \item \textbf{LesCorts:} A residential area near Camp Nou Football Stadium. Measurements at this location comprise 12 days (2019-01-12 to 2019-01-24).
    \item \textbf{PobleSec:} A residential and touristic area near the historic center, the exhibition centre and the port. Measurements at this location comprise 28 days (2018-02-05 to 2018-03-05).
    \item \textbf{ElBorn:} A touristic area in the downtown of the city. It is characterized by having a lot of shops and nightlife. Measurements at this location comprise 7 days (2018-03-28 to 2018-04-04).
\end{itemize}

As showed in~\cite{perifanis20235fed5g}, the dataset presents non Independent and Identically Distributed (non-IID) distibution, with the presence of three different skews, i.e. quantity, quality and temporality. This differentiation among respective Bias Sources (BS) is depicted in Fig.~\ref{fig:distribution}, indicating a challenging ML task.

\begin{figurehere}
\begin{center}
\includegraphics[width=2in]{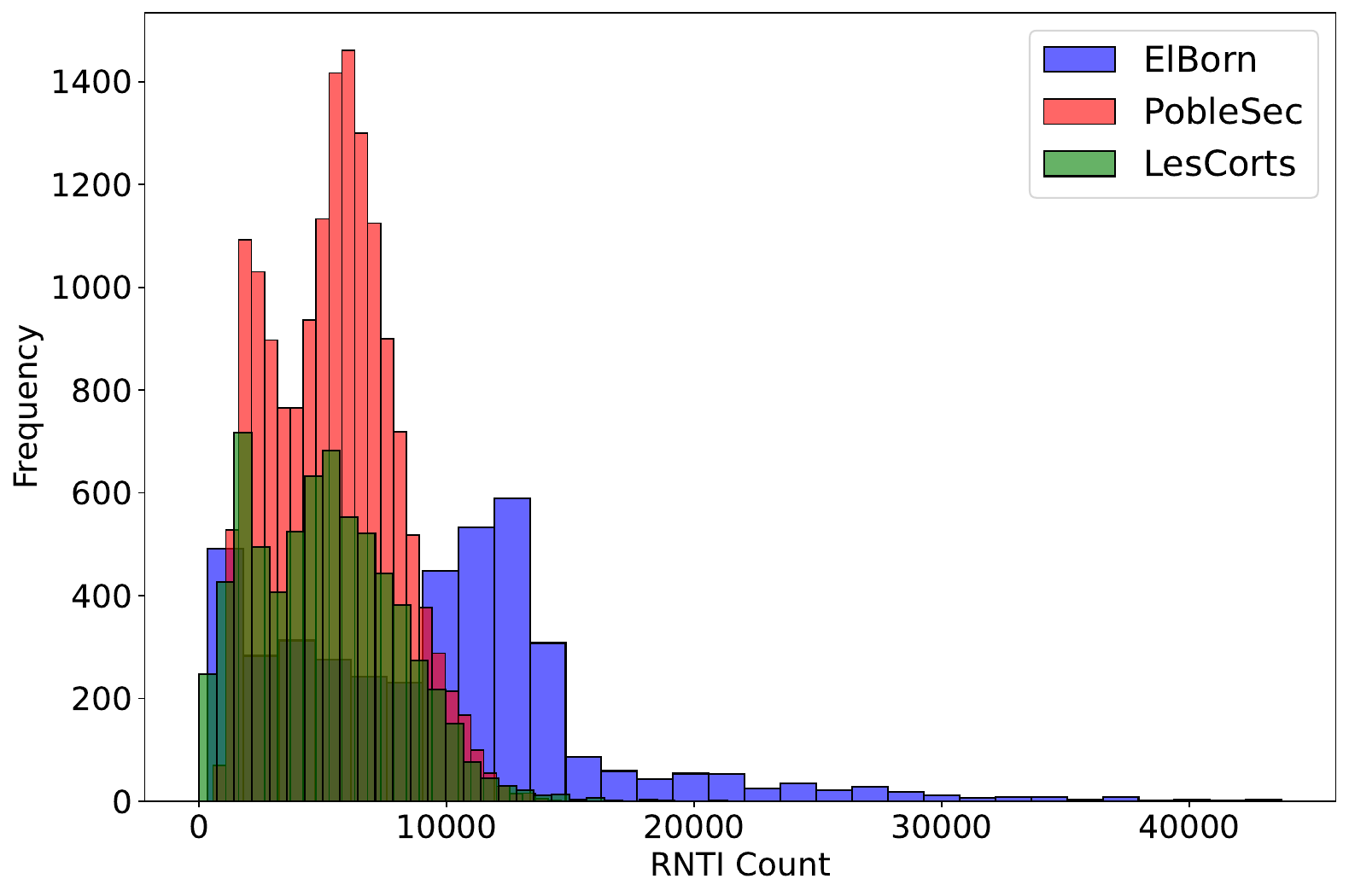}
\caption{Dataset Distribution ~\cite{Tsiolakis2024}}
\label{fig:distribution}
\end{center}
\end{figurehere}

\subsection{Problem Statement}
We formulate the problem in the context of time-series forecasting. Consider a dataset $\mathbf{X} \in \mathbb{R}^{N \times d}$, where $N$ is the number of observations in the dataset, and $d$ is the number of measurements recorded at each time step. In our context, these measurements include traffic-related metrics such as the uplink and downlink traffic and the number of connected devices at each time step. Given a window of $S$ past observations $\mathbf{X}_{t-S+1:t} = \{\mathbf{x}_{t-S+1}, \mathbf{x}_{t-S+2}, \ldots, \mathbf{x}_{t}\}$, where $\mathbf{x}_{i} \in \mathbb{R}^{d}$ is the measurement vector at time $i$, the goal is to predict the measurement vector $\mathbf{x}_{t+1} \in \mathbb{R}^{d'}$ at the next time step $t+1$, where $d' \leq d$ represents the dimensions of interest for forecasting.

The objective can be formalized as learning a function $f$ that maps a sequence of $S$ past observations to a future value:
\begin{equation}
    \hat{\mathbf{x}}_{t+1} = f(\mathbf{X}_{t-S+1:t}; \theta),
\end{equation}
where $\hat{\mathbf{x}}_{t+1} \in \mathbb{R}^{d'}$ is the predicted measurement vector for the next time step, and $\theta$ represents the parameters of the forecasting model. The model aims to minimize the forecasting error between the predicted measurements $\hat{\mathbf{x}}_{t+1}$ and the actual measurements $\mathbf{x}_{t+1}$.

\subsection{ML Models}
We evaluate the performance of several model architectures within our framework. Specifically, we first employ two traditional baseline learning models, i.e., MLP and CNN as benchmarks. To facilitate a comprehensive comparative analysis, we then selected several distinct Leaky-and-Integrate Fire (LIF) spiking neuron types:  the Lapicque's original Neuron, as well as its streamlined counterpart, the Leaky Model, alongside the Recurrent Leaky Neuron, the Synaptic Conductance-Based LIF Neuron Model, and the Alpha Neuron Model and a Reservoir Computing (RC), Echo State model. Specifically, we utilize the following model architectures:

\textbf{Multi-Layer Perceptron (MLP)}: 
    A MLP belongs to the class of feed-forward artificial neural networks (ANNs). MLP can not operate directly on matrix representations, thus requiring time-series to be flattened to one-dimensional vectors. In our case, we transform the $(S, d)$ two-dimensional array used as input into an one-dimensional vector to $(S \times d)$. In our architecture, we utilize a single hidden layer with 128 neurons.
    
\textbf{Convolutional Neural Network (CNN)}:
    CNNs are a type of deep neural networks, highly effective in analyzing data with a grid-like topology. In multivariate time-series, convolution operations can effectively utilize structural covariance. The implemented CNN takes a two-dimensional matrix of size $(S, d)$ as input and feeds it to four two-dimensional convolutional layers. The output is then propagated to a two-dimensional average pooling layer and finally to a fully connected layer. For the final layer, similar to MLP, we utilize a single hidden layer with 128 units.

\textbf{Reccurent Neural Network (RNN)}:
    RNNs are designed to capture temporal dependencies by maintaining a hidden state across sequential inputs. In our implementation, the RNN processes the time-series as an (S, d) sequence, where each step is passed through a recurrent layer with 128 hidden units, followed by a fully connected layer for prediction. 
    
\textbf{Lapicque Neuron (Original LIF Neuron)}:
    The Lapicque neuron implements the fundamental circuit approach, originating from biology. Ion channels of a physical neuron's membrane are modeled with a resistor and the membrane itself is modeled with a capacitor. We check the potential ($U$) in the capacitor, and if this potential reaches a predefined threshold, the neuron fires an output spike. We also choose a mechanism that is activated when a spike is fired to reset our neuron. Thus, the tunable parameters in this approach are the resistance ($R$), capacitance ($C$), threshold, the number of spikes for each input ($I_{in}$) and the reset mechanism. Formally, Lapicque neuron \cite{eshraghian2021snntorch} is modeled through the following function:
    \begin{equation}
    U[t+1] = (1 - \frac{1}{RC}) \cdot U[t] + \frac{1}{RC} \cdot I_{in}[t + 1] \cdot R,
    \end{equation}
    When the resulting value of $U$ is higher than the specified threshold, the neuron will fire a spike. 
    The threshold is a hyperparameter and it needs to be tuned to our problem to achieve, better predictive performance.
        
\textbf{Leaky Neuron}:
    The initial description of the LIF model, as described in~\cite{Gerum2021}, is distinguished by a broad set of hyperparameters. As the model is scaled to support a universal SNN architecture, managing these parameters becomes challenging. For this type of neuron, we use the following equation:
    \begin{equation}
        U[t + 1] = \beta \cdot U[t] + (1 - \beta) \cdot I_{in}[t + 1],
    \end{equation}
    where \(R = 1\) and \(\beta = (1 - \frac{1}{C})\), represeting a decay rate of the membrane potential. The spikes in the input increase the membrane potential, and when this value exceeds the threshold, a spike is fired in the output.

\textbf{Recurrent Leaky (RLeaky) Neuron}:
    This implementation, as discussed in \cite{eshraghian2021snntorch}, employs a logic similar to that of Leaky neuron, yet it uniquely scales the output spikes of a neuron by a factor of $V$ and reincorporates them into the input. Specifically, spikes produced by a layer of spiking neurons are forwarded to the following fully connected (or convolutional) layer, where they are combined with the incoming input before being processed by the spiking neuron model.
    
\textbf{Synaptic Based Neuron}:
    In the neuron models discussed previously, the premise has been that a voltage spike upon input instantaneously elevates the synaptic current, subsequently influencing the membrane potential. However, in practical scenarios, a spike initiates a progressive discharge of neurotransmitters from the pre-synaptic to the post-synaptic neuron. The synaptic conductance-based \cite{Bi1999syn} LIF model accommodates the slow temporal progression of the input current.
    Thus, the equation is transformed to:
    \begin{equation}
         U[t + 1] = \beta \cdot U[t] +  I_{syn}[t + 1], 
     \end{equation}
    where $I_{syn}[t + 1] = \alpha \cdot I_{syn}[t] +  I_{in}[t + 1]$, $\alpha$ is the decay rate of the synaptic current, and $\beta$ is the decay rate of the membrane potential.
    
\textbf{Alpha Neuron}:
    This model, as conceptualized in \cite{comsa2021alpha}, resembles a type of filter. Upon receiving an input spike, it is convolved with the filter, eliciting a change in the membrane potential. The configuration of the filter can be as simple as an exponential curve, or involve multiple exponentials. The attractiveness of these models lies in their ability to easily integrate a range of characteristics, such as refractoriness and threshold adjustment, by embedding these features within the design of the filter itself.

\textbf{Reservoir Computing Echo State}:
    Reservoir computing leverages a fixed, randomly initialized dynamic reservoir to project input data into a high-dimensional feature space. The Echo State Network (ESN) is a prominent implementation of this paradigm, particularly suited for sequential data processing. The reservoir consists of recurrently connected nodes, which transform the input time-series into a rich set of temporal features. Unlike traditional neural networks, the reservoir weights remain fixed, and only the output layer is trained, significantly reducing the computational complexity. In our implementation, the input data $(S, d)$ is mapped to the reservoir with a dimensionality of $N_r$ nodes. This dimensionality need to be tuned for our problem through a number of trials. The output is then computed as a linear combination of the reservoir states and passed to a fully connected layer for final predictions. This architecture facilitates efficient processing of multivariate time-series with minimal parameter tuning and low energy consumption. We selected Echo State Networks (ESNs) as a representative model of the reservoir computing paradigm due to their well-documented ability to handle temporal tasks with minimal training overhead. ESNs require training only the output layer, resulting in significantly reduced computational and energy demands—an important factor in our study of sustainable AI. Their simplicity and reproducibility also make them an ideal baseline for comparison. While other models within reservoir computing (e.g., Liquid State Machines) could be considered, our focus was to benchmark a widely adopted and computationally efficient architecture.
    In our experiments we use 128 reservoir neuron with fixed weights and only train the weights of the output layer.

\subsection{Evaluation Metrics}
The primary metric used to quantify prediction error for the time-series forecasting task is the normalized root mean squared error (\(\mathrm{NRMSE}\)), defined as $\mathrm{NRMSE} = \frac{1}{\overline{x}}\sqrt{\frac{\sum_{t=1}^{T}\left(\sum_{i=1}^{d'} \left(\hat{x}_{t,i} - x_{t,i} \right)^2\right)}{Td'}}$,
where \(\overline{x}\) represents the mean of the target observations, computed as $\overline{x} = \frac{1}{Td'}\sum_{t=1}^{T}\left(\sum_{i=1}^{d'} x_{t,i}\right)$,
and \(\hat{x}_{t,i}\) and \(x_{t,i}\) denote the predicted and actual values at time \(t\) for feature \(i\), \(T\) is the total number of time steps, and \(d'\) is the number of features.

In addition to \(\mathrm{NRMSE}\), we also employ other common error metrics to evaluate model performance: The Mean Absolute Error $\mathrm{MAE}=\frac{1}{Td'} \sum_{t=1}^{T} \sum_{i=1}^{d'} \left| \hat{x}_{t,i} - x_{t,i} \right|$, the Root Mean Square Error $\mathrm{RMSE}=\sqrt{\frac{1}{Td'} \sum_{t=1}^{T} \sum_{i=1}^{d'} \left(\hat{x}_{t,i} - x_{t,i}\right)^2}$ and the Mean Square Error $\mathrm{MSE}=\frac{1}{Td'} \sum_{t=1}^{T} \sum_{i=1}^{d'} \left(\hat{x}_{t,i} - x_{t,i}\right)^2$.

Beside the accuracy of the implemented algorithms, we also utilize the Sustainability Indicator $\mathcal{S}$ introduced in~\cite{perifanis2023towards} to assess the resulting trade-off between the prediction error and the energy consumption. $\mathcal{S}$ enables a fair comparison among different model architectures and can be calculated using the following formula:
\begin{equation}
    \label{eq:sus}
    \mathcal{S} = (1 + E_{\text{Val}})^a \times (1 + C_{\text{Tr}})^b \times (1 + D)^c ,
\end{equation} 
where $E_{\text{Val}}$ is the validation error (in this work, we consider NMRSE),  $C_{\text{Tr}}$ represents the total energy consumed for model training in Watt Hours (Wh) and $D$ represents the size of the data exchanged during training. In centralized learning, $D$ corresponds to the total dataset size transferred to a central server. In contrast, in federated learning, $D$ denotes the size of the model weights exchanged between clients and the server, as only model updates are communicated rather than raw data. The use of this parameter, instead of direct power consumption during communication, is explained in~\cite{perifanis2023towards}. This choice is primarily based on the fact that, although direct communication costs are often considered negligible ~\cite{Guerra2023} , a broader sustainability analysis should also account for indirect factors—such as throughput and bandwidth requirements—which can impact the system’s overall energy footprint (e.g., increased cooling costs of data centers). The exponents
denote the importance of each value, with $a + b + c = 1$.
The lower the value of $S$, the better the tradeoff between computational efficiency and predictive accuracy.

\section{Experimental Setup} \label{sec:experiments}

In our SNN models, both the input and output layers utilize the Subtract-Reset Mechanism. This means that when a neuron's membrane potential exceeds the threshold, a spike is emitted, and the membrane potential is reduced by the threshold value (\( V_{\text{mem}} = V_{\text{mem}} - \text{threshold} \)).

In the output layer, however, neurons lack a reset mechanism—a deliberate design choice to better address the requirements of the time-series forecasting task. This configuration ensures that the predicted output corresponds directly to the membrane potential, avoiding distortions caused by resets. Additionally, input spikes are encoded directly from the input data at specific timesteps, enabling the network to effectively convert non-spiking data into a format compatible with spiking mechanisms. This approach enhances the model's ability to process time-series data within the spiking neural network framework. To ensure consistency, we employed the same dataset throughout all experiments. Moreover, we set $a=b=c=0.33$ to assign equal weight to power consumption, predictive performance, and size parameter in the index.

\begin{table*}
\tbl{Best Hyperparameter Configuration for Each Model after Tuning via Optuna Python Library \label{tab:model_hyperparam}}
{\begin{tabular}{lccccc}
\hline
\textbf{Model} & \textbf{Hidden Neurons} & \textbf{Threshold} & \textbf{Learning Rate} & \textbf{Batch Size} \\ \hline
Lapicque Neuron              & 256 & 1.6 & $1.09 \times 10^{-4}$ & 130  \\
Leaky Neuron                 & 96  & 1.7 & $2.39 \times 10^{-4}$ & 64  \\
Recurrent Leaky Neuron       & 256 & 2.0 & $5.85 \times 10^{-4}$ & 32  \\
Synaptic Based Neuron        & 128 & 1.7 & $1.07 \times 10^{-4}$ & 128 \\
Alpha Neuron                 & 256 & 2.0 & $5.85 \times 10^{-4}$ & 32  \\
Multi-Layer Perceptron       & 128 & --  & $ 10^{-3}$            & 128 \\
Convolutional Neural Network & 128 & --  & $ 10^{-3}$            & 128 \\ \hline
\end{tabular}}
\end{table*}

For implementation, we utilized the PyTorch library~\cite{paszke2019pytorch} for traditional models such as MLPs and CNNs. The SNN models were implemented using the snnTorch library~\cite{eshraghian2021snntorch}, which is tailored for gradient-based learning and provides pre-designed spiking neuron models that seamlessly integrate as recurrent activation units. 

To ensure the validity and fairness of our comparative analysis, we employed Optuna Python Library ~\cite{optuna_2019}, that uses a Tree-structured Parzen Estimator (TPE) to tune key parameters in the centralized learning setup. Specifically, we optimized the number of hidden-layer neurons, threshold values, learning rate, batch size. The best-performing hyperparameter configurations identified through this process are summarized in Table ~\ref{tab:model_hyperparam}. These configurations were subsequently applied to the federated learning setup to ensure consistency and comparability across different settings. We employed 150 epochs for the centralized learning, and 50 Federated Rounds with 3 Local epochs in the federated setup to ensure that the dataset was accessed the same number of times. Additionally, we incorporated an early stopping mechanism during training of both learning settings, halting the process if no improvement in the loss function was observed over 50 consecutive epochs. This not only prevented overfitting but also contributed to energy efficiency by reducing unnecessary computations.

Energy consumption during the training process was measured using the CarbonTracker Python module~\cite{anthony2020carbontracker}. CarbonTracker is a tool designed to estimate and monitor the carbon emissions associated with computational processes, helping organizations and researchers assess and reduce their environmental impact. The experiments were conducted on a Windows 11 workstation equipped with an AMD Ryzen 7 6800HS CPU and 16 GB of memory. \footnote{The source code used in this paper is available at \href{https://github.com/vperifan/Federated-Time-Series-Forecasting}{https://github.com/vperifan/Federated-Time-Series-Forecasting}, accessed on May 10 2025.}

\section{Experimental Evaluation} \label{sec:evaluation}

\subsection{Predictive Performance}
\subsubsection{Modeling of Input Data}
Selecting the correct number of spiking timesteps is crucial for effectively encoding input data in SNN models. The number of timesteps determines how the input data is represented as spikes, influencing the network's ability to process and learn from the encoded information. Using too few timesteps may result in a loss of critical details during the encoding process, leading to poor model performance. On the other hand, an excessive number of timesteps can overcomplicate the encoding, increasing computational costs and energy consumption without improving the predictive capacity of the model. This challenge highlights the importance of fine-tuning the timestep configuration to achieve an efficient and accurate representation of the input data.

In our experiments, we investigated how varying the number of timesteps used to encode input data impacts the predictive performance of the models. Specifically, we tested configurations with 1, 7, 10, 50, 70, and 100 timesteps to identify the optimal setup for maximizing accuracy.

As shown in Figure~\ref{fig:snn_timesteps}, the results indicate significant variations in performance depending on the chosen timestep length. Optimal performance was achieved at 10 timesteps for some models and 7 timesteps for others. Beyond these values, increasing the number of timesteps did not improve accuracy and, in some cases, resulted in performance degradation. These findings highlight the importance of selecting an appropriate timestep configuration to balance information completeness, computational efficiency, and predictive accuracy.

\begin{figurehere}
\begin{center}
\includegraphics[width=0.8\columnwidth]{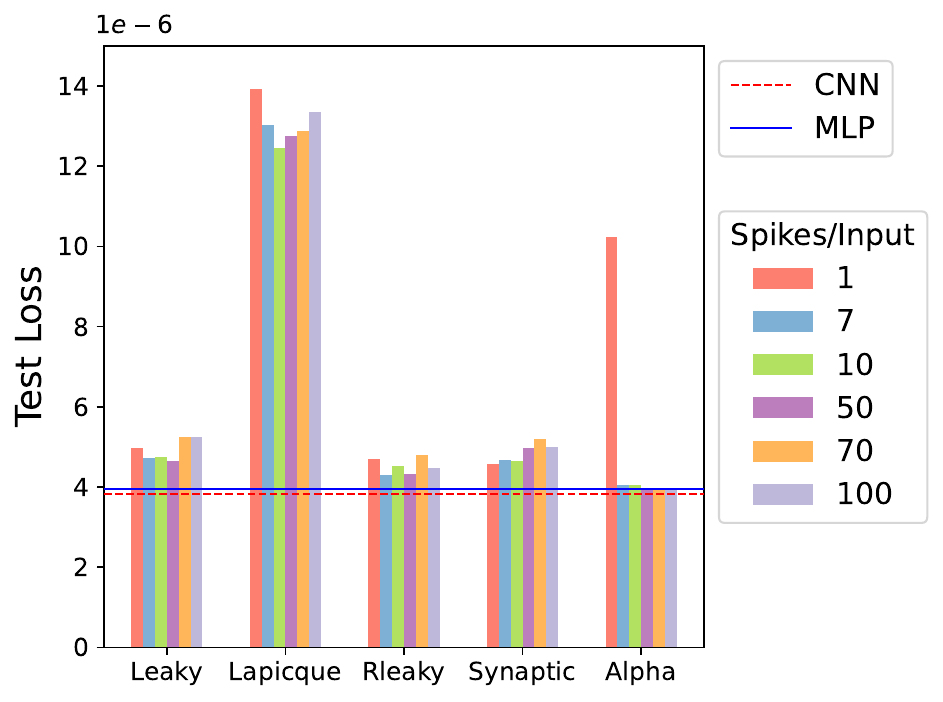}
\caption{Test Loss for different number of spikes per input}
\label{fig:snn_timesteps}
\end{center}
\end{figurehere}

Based on these observations, we selected 7 and 10 timesteps for our examples, as they demonstrated the best performance. This choice minimizes the increase in computational complexity, keeping the calculations as simple as possible to reduce power consumption, while maintaining predictive performance at an acceptable level.

\subsubsection{Centralized Learning}
\begin{figure*}[ht]
\center
\includegraphics[width=0.8\textwidth]{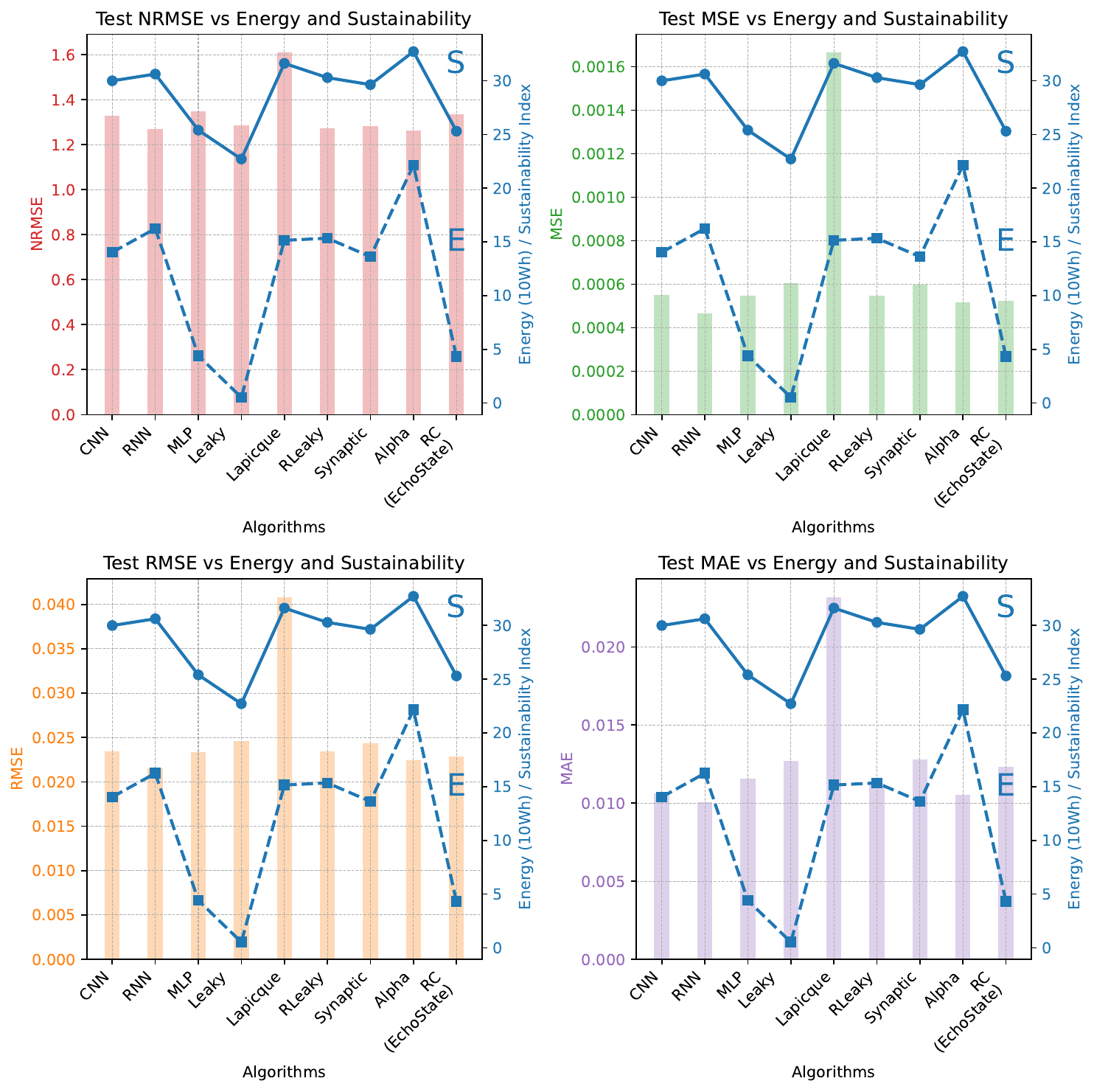}
\caption{Energy vs Prediction 7 Timesteps}
\label{fig:Centralized7timesteps}
\end{figure*}

The evaluation of model performance and sustainability across centralized configurations reveals critical insights into the trade-offs between predictive accuracy, energy efficiency, and overall sustainability. Figure \ref{fig:Centralized7timesteps} present the performance of various models using metrics such as NRMSE, MSE, RMSE, and MAE. The results show that model performance varies significantly depending on the selected metric. To ensure consistency and comparability, NRMSE was chosen as the primary metric, as it normalizes errors relative to the data scale, providing a more realistic and interpretable estimation. 

From the data in Tables \ref{tab1} and \ref{tab2}, as well as the referenced figures, we analyzed the performance of the models for 7 and 10 timesteps. The findings reveal that the Leaky and RLeaky architectures are the models that perform better with 7 timesteps. In contrast, all other models achieve superior performance and lower error rates with 10 timesteps, highlighting the importance of selecting an optimal timestep configuration for each model to balance computational efficiency and predictive accuracy.

Regarding sustainability, an initial inspection of the Sustainability Index suggests that the relationship between energy efficiency and predictive performance is not straightforward. Reducing model complexity often leads to lower power consumption but can result in higher error rates, which negatively impacts the Sustainability Index (where a lower value of S indicates greater sustainability). Ultimately, the Sustainability Index reflects the balance between energy efficiency and performance, and the critical question becomes how much power consumption can be reduced without a significant sacrifice in predictive accuracy. Among the centralized configurations, the Leaky Neuron stands out as the most sustainable model during the training process. Despite its higher prediction error compared to RNN, its predictive capacity compared to MLP, CNN, and its remarkably low power consumption compensates for this limitation, resulting in the lowest S. This makes the Leaky Neuron the best overall performer in terms of the examined trade-off.

In contrast, the Alpha Neuron  with 7 timesteps demonstrates the best predictive accuracy among all centralized models, indicating its ability to effectively handle the task. However, this performance comes at the expense of significantly higher power consumption, as the model operates as a computationally intensive filter. Consequently, it has the worst S among all evaluated models.
The Lapicque Neuron exhibits the poorest performance, with the highest RMSE and a power consumption higher than both the MLP and CNN models. This combination of low accuracy and high energy demand results in the lowest S, making the Lapicque Neuron the least favorable choice for centralized configurations. The RLeaky model presents a particularly intriguing case. While it achieved its best performance at 10 timesteps with improved predictions and a lower NRMSE, this came at the cost of significantly higher power consumption during training. This trade-off highlights the challenge of balancing accuracy and energy efficiency, making the RLeaky model less suitable for applications where sustainability is a critical consideration. The Echo State model demonstrates a better balance between accuracy and power consumption. Ranking second in terms of sustainability, it outperforms models such as the Lapicque Neuron, MLP, and CNN. Its architecture leverages energy efficiency without substantial performance compromises, making it a promising alternative for resource-constrained applications. Finally, the Synaptic model, which aims to replicate the complex functions of brain cells, demonstrates significant drawbacks in terms of resource requirements. It consumes substantially more power and produces higher NRMSE compared to the MLP and CNN. This result suggests that the additional biological realism in the Synaptic model comes at the expense of computational efficiency and predictive accuracy, raising concerns about its practicality in real-world applications.

Collectively, these findings highlight that certain bio-inspired models are capable of reducing power consumption, which in turn contributes to lower carbon emissions.

\begin{table*}
\tbl{Centralized Cumulative results on NRMSE, Energy Consumption and Sustainability of considered models.}
{\begin{tabular}{c|cc|cc|cc}
\hline
Model    & \multicolumn{2}{c|}{NRMSE}           & \multicolumn{2}{c|}{Consumption (Wh)} & \multicolumn{2}{c}{$\mathcal{S}$}    \\ \hline
MLP      & \multicolumn{2}{c|}{1.3465}          & \multicolumn{2}{c|}{\textbf{0.44}}             & \multicolumn{2}{c}{25.398}           \\
RNN      & \multicolumn{2}{c|}{\textbf{1.2693}}          & \multicolumn{2}{c|}{1.624}             & \multicolumn{2}{c}{30.62}           \\
CNN      & \multicolumn{2}{c|}{1.3272}          & \multicolumn{2}{c|}{1.403}            & \multicolumn{2}{c}{29.992}           \\ \hline
Echo State (RC)      & \multicolumn{2}{c|}{1.3343}          & \multicolumn{2}{c|}{0.431}            & \multicolumn{2}{c}{\textbf{25.302}}           \\ \hline
 &
  \multicolumn{1}{c|}{7 spikes/input} &
  10 spikes/input &
  \multicolumn{1}{c|}{7 spikes/input} &
  10 spikes/input &
  \multicolumn{1}{c|}{7 spikes/input} &
  10 spikes/input \\ \hline
Leaky &
  \multicolumn{1}{c|}{1.2870} &
  1.2864 &
  \multicolumn{1}{c|}{\textbf{0.0531}} &
  \textbf{0.0713} &
  \multicolumn{1}{c|}{\textbf{22.713}} &
  \textbf{22.839} \\
Lapicque & \multicolumn{1}{c|}{1.6099} & 1.6358 & \multicolumn{1}{c|}{1.5143}  & 1.813  & \multicolumn{1}{c|}{31.6174} & 32.9177 \\
RLeaky   & \multicolumn{1}{c|}{1.2732} & 1.2678 & \multicolumn{1}{c|}{1.5343}  & 1.862  & \multicolumn{1}{c|}{30.287} & 31.503 \\
Synaptic & \multicolumn{1}{c|}{1.2832} & 1.3132   & \multicolumn{1}{c|}{1.3623}  & 2.543   & \multicolumn{1}{c|}{29.6362} & 34.024 \\
Alpha &
  \multicolumn{1}{c|}{\textbf{1.2639}} &
  \textbf{1.2872} &
  \multicolumn{1}{c|}{2.2141} &
  3.234 &
  \multicolumn{1}{c|}{32.713} &
  35.95 \\ 
  \hline
\end{tabular}
\label{tab1}}
\end{table*}

\subsubsection{Federated Learning}
\begin{figure*}[ht]
\center
\includegraphics[width=0.8\textwidth]{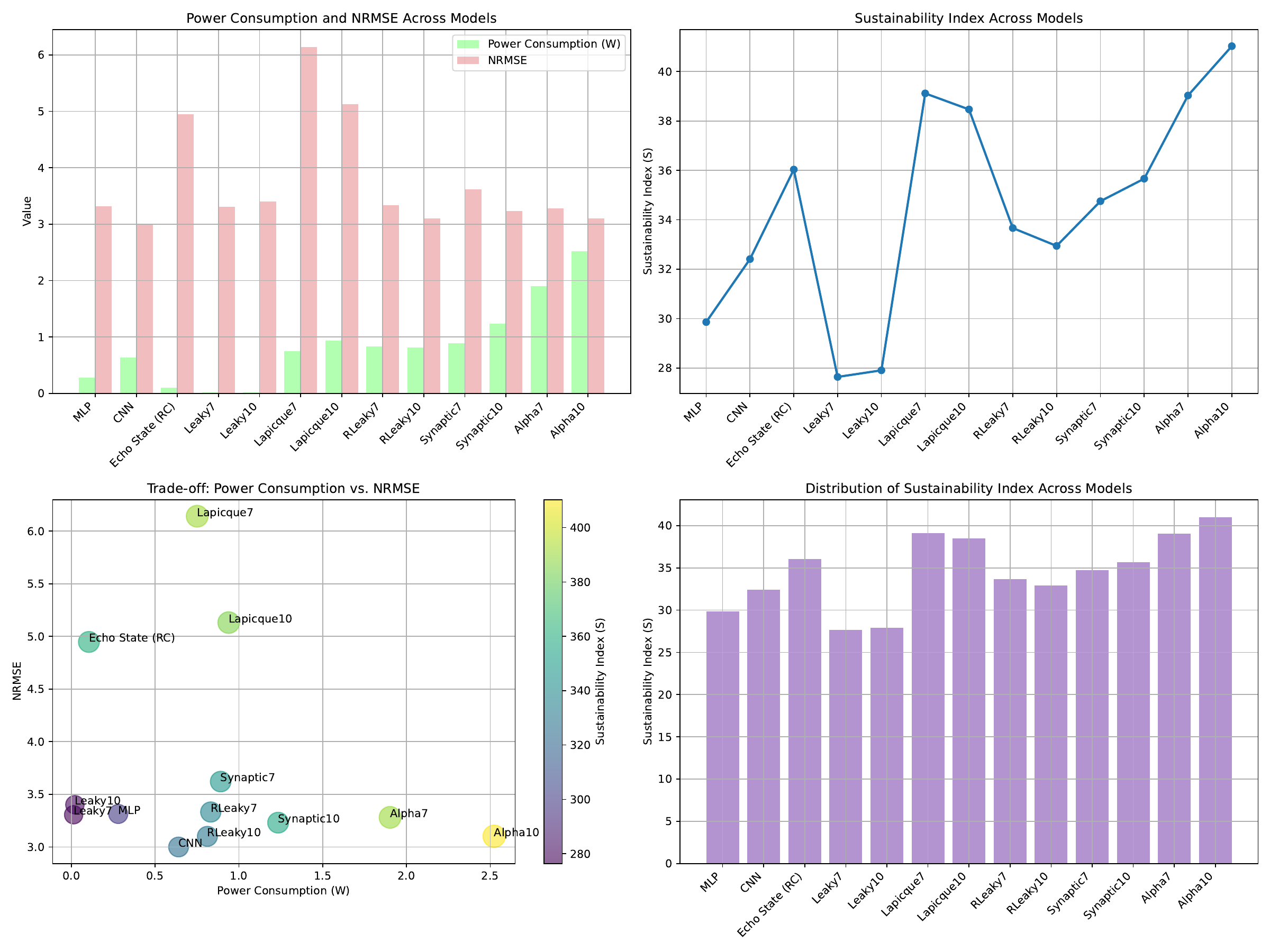}
\caption{Federated Learning Results (i) the Power consumption (Green) and the NRMSE (Pink) across different models in federated learning. (ii) the sustainability index (S) for the federated models. (iii) The trade-off between the NRMSE and the Power Consumption, the color of each dot represents the sustainability index. (iv) The sustainability index across different models, starting the y axis from 0.}
\label{fig:fedResults}
\end{figure*}

The graphs in Figure~\ref{fig:fedResults} provide a comprehensive comparison of energy consumption, predictive performance, and sustainability of federated learning configurations across various machine learning models. In those graphs the number in the model name means the number of timesteps, e.g. the Leaky7 means the Leaky model with 7 timesteps.

The first graph illustrates the energy consumption of the models under the two paradigms, 7 and 10 timesteps, giving us a point to understand how in this federated setup the power consumption, reflects the predictive capacity. The second graph compares the Normalized Root Mean Square Error (NRMSE) of the models, offering insight into their predictive accuracy. The third graph examines the trade-off between energy consumption and NRMSE, visualizing the balance between resource efficiency and predictive accuracy. This trade-off is critical for assessing the suitability of different learning paradigms in energy-constrained environments, such as edge devices or mobile applications. The final graph compares the S across models, integrating both energy efficiency and predictive performance. Federated models, however, show varying levels of sustainability depending on the architecture, emphasizing the importance of selecting the appropriate paradigm based on application-specific trade-offs among energy, accuracy, and sustainability.

Additionally, from Table \ref{tab2}, we observe that in the federated setup, the RNN model achieves the most accurate predictions. Its performance is higher comparable to that to the next better neural network architecture with Alpha neurons. Alpha outperform the traditional MLP and CNN model. Furthermore, predictions made with Lapicque neurons are significantly less accurate, with an NRMSE approximately twice as high as that of the CNN model. The Leaky neuron also performs competitively in the federated setup, achieving a lower NRMSE than the MLP, although its error remains slightly higher than that of the CNN and RNN.

\begin{table*}
\tbl{Federated Cumulative results on NRMSE, Energy Consumption and Sustainability of considered models.}
{\begin{tabular}{c|cc|cc|cc}
\hline
Model    & \multicolumn{2}{c|}{NRMSE}           & \multicolumn{2}{c|}{Consumption (Wh)} & \multicolumn{2}{c}{$\mathcal{S}$}    \\ \hline
MLP      & \multicolumn{2}{c|}{3.3134}          & \multicolumn{2}{c|}{0.28}             & \multicolumn{2}{c}{29.86}           \\
RNN      & \multicolumn{2}{c|}{\textbf{1.529}}          & \multicolumn{2}{c|}{0.832}             & \multicolumn{2}{c}{\textbf{28.186}}           \\
CNN      & \multicolumn{2}{c|}{2.9999}          & \multicolumn{2}{c|}{0.64}            & \multicolumn{2}{c}{32.410}           \\ \hline
Echo State (RC)      & \multicolumn{2}{c|}{4.8372}          & \multicolumn{2}{c|}{\textbf{0.096}}            & \multicolumn{2}{c}{31.3536}           \\ \hline
 &
  \multicolumn{1}{c|}{7 spikes/input} &
  10 spikes/input &
  \multicolumn{1}{c|}{7 spikes/input} &
  10 spikes/input &
  \multicolumn{1}{c|}{7 spikes/input} &
  10 spikes/input \\ \hline
Leaky &
  \multicolumn{1}{c|}{3.3019} &
  3.3854 &
  \multicolumn{1}{c|}{\textbf{0.0125}} &
  \textbf{0.0207} &
  \multicolumn{1}{c|}{\textbf{27.6385}} &
  \textbf{27.8677} \\
Lapicque & \multicolumn{1}{c|}{5.869} & 5.127 & \multicolumn{1}{c|}{0.728}  & 0.895  & \multicolumn{1}{c|}{38.4474} & 38.1685 \\
RLeaky   & \multicolumn{1}{c|}{3.289} & 3.046 & \multicolumn{1}{c|}{0.842}  & 0.890  & \multicolumn{1}{c|}{33.6144} & 33.2547 \\
Synaptic & \multicolumn{1}{c|}{3.598} & 3.126  & \multicolumn{1}{c|}{0.945}  & 1.165   & \multicolumn{1}{c|}{35.0182} & 35.0049 \\
Alpha &
  \multicolumn{1}{c|}{\textbf{3.2765}} &
  \textbf{3.0984} &
  \multicolumn{1}{c|}{1.765} &
  2.265 &
  \multicolumn{1}{c|}{38.3991} &
  39.9988 \\ \hline
\end{tabular}
\label{tab2}}
\end{table*}

\subsubsection{Federated vs. Centralized}

\begin{figure*}[ht]
\center
\includegraphics[width=0.8\textwidth]{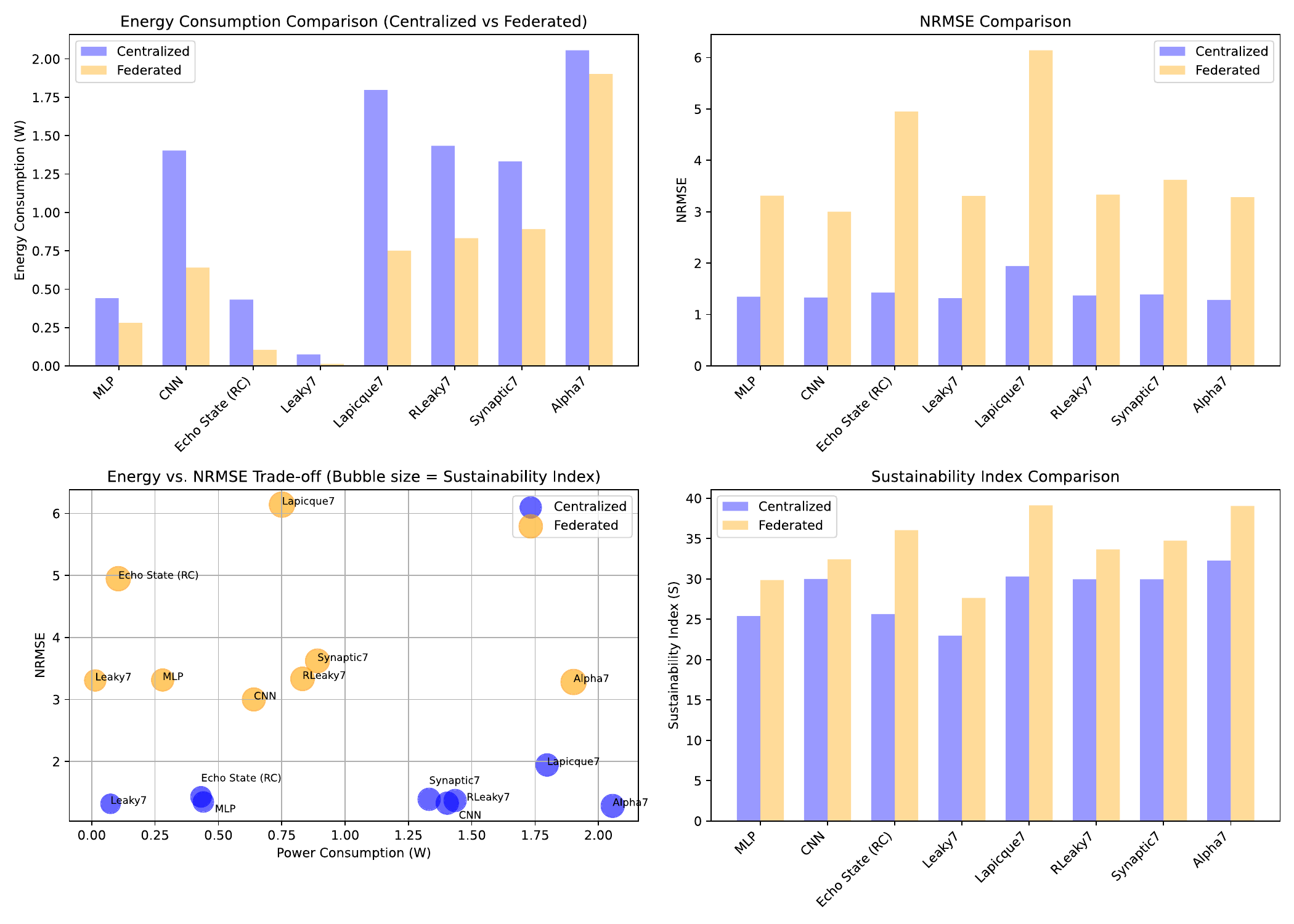}
\caption{Federated vs Centralized Learning 7 timesteps. (i) Comparison of the Power Consumption of the centralized (Purple) and the federated (yellow) (ii)  Comparison of the NRMSE of the centralized (Purple) and the federated (yellow) (iii) Energy and NRMSE trade off, the color indicates the federated or the centralized configuration, and the radius of the circle the magnitude of the sustainabilitay index. (iv)  Comparison of the Sustainability index (S) of the centralized (Purple) and the federated (yellow).}
\label{fig:fedrVScentr7}
\end{figure*}

From Figure \ref{fig:fedrVScentr7}, which comprises four visualizations, we observe that in our simulation, federated power consumption is significantly lower compared to centralized configurations. However, this reduction in power consumption comes at the cost of a noticeably higher NRMSE. Furthermore, for the federated configuration, we find that for different models and configurations the performance preserves a volatility with 7 and 10 timesteps. This observation underscores the importance of selecting an optimal timestep configuration to balance computational complexity, energy efficiency, and predictive accuracy.

The first graph of figure \ref{fig:fedrVScentr7} illustrates the energy consumption of centralized and federated models. Federated configurations demonstrate consistently lower energy consumption, whereas centralized models exhibit higher energy usage. The second graph compares the predictive performance of the models using the Normalized Root Mean Square Error (NRMSE) metric. Centralized models achieve superior predictions, reflected in their lower NRMSE values. Federated models, while showing higher NRMSE values, maintain reasonable performance levels, making them viable for applications where energy restrictions exists and power consumption are prioritized over predictive precision. The third graph visualizes the trade-off between energy consumption and predictive accuracy, with a scatter plot of energy consumption versus NRMSE. Points closer to the origin (0, 0) represent models that achieve both high resource efficiency and low predictive error. Centralized models tend to cluster near the lower end of the NRMSE axis, indicating higher accuracy, but many of them exhibit higher values along the power consumption axis. In contrast, federated models demonstrate a clearer trade-off: most show lower energy consumption but at the cost of slightly reduced predictive accuracy. 

The final graph compares the sustainability index (S), which integrates energy efficiency and predictive performance. Centralized models achieve better sustainability scores due to their higher predictive capacity. Federated models exhibit varying levels of sustainability depending on the architecture and configuration; however, they generally underperform compared to centralized configurations in terms of sustainability.

Together, these findings emphasize the strengths and limitations of centralized and federated learning approaches. They also highlight the importance of carefully selecting configurations, such as timestep lengths and model architectures, to achieve an optimal balance between energy efficiency, predictive performance, and sustainability in diverse operational contexts.

It is evident from this analysis that the most sustainable models across all experiments are the centralized Leaky models with 7 timesteps and 10 timesteps. Among these, the centralized Leaky model with 7 timesteps is the most sustainable, achieving an S value of 22.713. In other words, this model achieves the best overall trade-off among energy consumption, predictive accuracy, and communication cost. However, this outcome can vary depending on the specific priorities of a given application. By adjusting the exponents in the Sustainability Index, one can emphasize particular aspects of the trade-off to better align with application-specific requirements.

\section{Discussion and Open Issues} \label{sec:discussion}

We observe significant variations in power consumption across models during training. For example, the Alpha Neuron architecture for SNNs achieves lower prediction errors but at the cost of sustainability, due to higher power consumption. Notably, it demonstrates the best predictive capacity in our centralized training experiments. These findings highlight that there is no universal rule ensuring bio-inspired models inherently consume less power while maintaining low predictive error. Consequently, a clear trade-off cannot be universally established; each case must instead be analyzed individually.

We do, however, observe a general trend: increased computational complexity tends to result in higher power consumption alongside improved predictive performance. The challenge, therefore, is to select a model with a level of complexity that balances acceptable prediction error with manageable energy demands, tailored to the needs of a specific application. 

At the same time, we must acknowledge the limitations of current models in accurately simulating brain-like learning mechanisms. While the Alpha neuron achieves better predictive capacity by incorporating spiking elements that mimic brain functionality, its internal structure resembles a filter more than the biological structure of a neuron.

Meanwhile, the Synaptic neuron, which more closely resembles the functionality of brain cells, exhibits higher error rates compared to classic models but performs better than the Alpha neuron. Additionally, the Synaptic neuron consumes significantly more power than classic models, though less than the Alpha neuron. The increased power consumption of the Synaptic neuron arises from its attempt to simulate biological processes more accurately, resulting in more complex computations. This suggests there may be a limit to how closely we should model brain functionality. In our case, the more biologically accurate Synaptic neuron produced more erroneous results compared to the Leaky neuron, a simplified, classic Leaky-Integrate-and-Fire neuron.

These observations raise two key points: First, unlike the brain’s specialized cells, our models run on general-purpose hardware, highlighting the need to evaluate them on neuromorphic systems that better reflect their design and provide more accurate power measurements. Second, brain-inspired models may not naturally align with tasks like numerical prediction, so their strengths should be assessed in domains that better match biological processing capabilities.

In our experiments, centralized learning most outperformed federated learning in terms of predictive accuracy and the energy-prediction trade-off. For example, the centralized Echo State Network (ESN) achieved an NRMSE of 1.3343 and consumed 0.431 Wh, resulting in a sustainability index S = 25.302. In contrast, the federated ESN yielded a much higher NRMSE of 4.8372 despite consuming less energy (0.096 Wh), with a significantly worse sustainability index S = 31.3536. Similarly, the centralized Leaky model (7 timesteps) achieved the best overall sustainability (S = 22.713) with NRMSE = 1.2870, outperforming its federated counterpart (S = 27.6385, NRMSE = 3.3019) despite slightly higher energy usage (0.0531 Wh vs. 0.0125 Wh). The RNN model also demonstrates strong predictive performance on our time-series data in both centralized and federated settings, confirming its effectiveness for temporal tasks. However, its high energy consumption limits its suitability to resource-rich environments. Notably, the federated RNN achieved an NRMSE = 1.529, outperforming all other models in the federated setting.

Finally, it is worth highlighting some key aspects regarding the sustainability analysis. 

In the context of this work, federated learning proved more energy-efficient during the model training phase. A possible explanation of this is the early stopping mechanism applied at each client, which led to faster model convergence and reduced local computations in the decentralized setting.

Another important sustainability factor is data transmission. As previously discussed, federated learning requires only minimal data exchange—typically a few kilobytes of model weights—thereby improving the overall sustainability index.

Despite these advantages, federated models consistently underperformed in terms of prediction error across all experiments. This performance gap contributes to centralized training achieving a better overall Sustainability Index, as the metric also accounts for model accuracy.

Potential future directions include exploring hybrid architectures that combine SNNs with conventional RNNs and ESNs, as well as a variety of ensembling strategies—from simple vote-based schemes to component-level integrations—to assess their trade-offs between predictive performance and power efficiency within a sustainability framework. However, these approaches require careful consideration, as simply executing multiple models and aggregating their outputs can substantially increase power consumption and undermine sustainability goals. A very interesting future work will be to experiment with different encoding techniques of the input to timesteps.
Also, we aim to experiment with alternative architectures for Reservoir Computing, as well as explore various nature-inspired optimization algorithms with a potential to reduce the power consumption —such Intelligent Water Drops as it described in the Siddique et al. ~\cite{Siddique2014waterdrop}, Bacterial Foraging as it described in Wand et al. ~\cite{Wang2018}, and Spiral dynamics algorithms as described in ~\cite{siddique2014spiral} potential enhancements in the next phase of our research.

\section{Conclusion} \label{sec:conclusion}
In conclusion, bio-inspired models such as Leaky and Echo State demonstrate significant potential for promoting sustainable machine learning. Their inherent energy efficiency and scalability position them as strong candidates for a wide range of applications, particularly in resource-constrained environments. With further adaptations and optimizations, these models can be refined to achieve superior performance in specific tasks, enhancing their practicality for real-world deployment. Within the centralized and decentralized configuration, our findings highlight the Leaky model as the most sustainable option, offering a compelling solution for energy-efficient machine learning.

\bibliographystyle{ws-ijns}
\bibliography{main}

\end{multicols}
\end{document}